\documentclass{article}

\PassOptionsToPackage{numbers}{natbib}
\usepackage[final]{nips_2017}

\usepackage[utf8]{inputenc} % allow utf-8 input
\usepackage[T1]{fontenc}    % use 8-bit T1 fonts
\usepackage{hyperref}       % hyperlinks
\usepackage{url}            % simple URL typesetting
\usepackage{booktabs}       % professional-quality tables
\usepackage{amsfonts}       % blackboard math symbols
\usepackage{nicefrac}       % compact symbols for 1/2, etc.
\usepackage{microtype}      % 
\usepackage{times}
\usepackage{color,soul}
\usepackage{amsmath,amssymb}
\usepackage{bm}
\usepackage{bbm}
\usepackage{float}
\usepackage{graphicx}
\usepackage{subfig}
\usepackage{ctable}
\usepackage{multirow}
\usepackage{adjustbox}
\usepackage{rotating}
\usepackage{setspace}% http://ctan.org/pkg/setspace
\usepackage{lipsum}%http://ctan.org/pkg/lipsum

\newcolumntype{K}[1]{>{\centering\arraybackslash}p{#1}}
\title{Obtaining Accurate Probabilistic Causal Inference by Post-Processing Calibration}

\author{Fattaneh Jabbari$^1$, Mahdi Pakdaman Naeini$^2$, Gregory F. Cooper$^{1,3}$
      \\
      $^1$Intelligent Systems Program, University of Pittsburgh,
      Pittsburgh, PA\\
      $^2$Paulson School of Engineering and Applied Sciences, Harvard University, Cambridge, MA\\
      $^3$Department of Biomedical Informatics, University of Pittsburgh,
      Pittsburgh, PA\\
      \texttt{fattaneh.j@pitt.edu}, \texttt{pakdaman@seas.harvard.edu}, \texttt{gfc@pitt.edu}
      }

\begin{document} 
\maketitle 
\begin{abstract}
Discovery of an accurate causal Bayesian network structure from observational data can be useful in many areas of science. Often the discoveries are made under uncertainty, which can be expressed as probabilities. To guide the use of such discoveries, including directing further investigation, it is important that those probabilities be well-calibrated. In this paper, we introduce a novel framework to derive calibrated probabilities of causal relationships from observational data. The framework consists of three components: (1) an approximate method for generating initial probability estimates of the edge types for each pair of variables, (2) the availability of a relatively small number of the causal relationships in the network for which the truth status is known, which we call a \textit{calibration training set}, and (3) a calibration method for using the approximate probability estimates and the calibration training set to generate calibrated probabilities for the many remaining pairs of variables. We also introduce a new calibration method based on a shallow neural network. Our experiments on simulated data support that the proposed approach improves the calibration of causal edge predictions. The results also support that the approach often improves the precision and recall of predictions.
\end{abstract}
\vspace{-2 mm}
\section{Introduction}
\vspace{-1 mm}
Much of science consists of discovering and modeling causal relationships in nature. Increasingly, scientists have available multiple complex data and a large number of samples, each of which has an enormous number of measurements recorded, thanks to rapid advancements in sophisticated measurement technology, where this data are often purely observational. In past 25 years, there has been tremendous progress in developing computational methods for discovering causal knowledge from observational data \cite{glymour1999computation,spirtes2000causation,pearl2003causality,spirtes2010introduction,illari2011causality}. A primary use of such methods is to analyze observational data to generate novel causal hypotheses that are likely to be correct when subjected to experimental validation; such an approach can significantly increase the efficiency of causal discovery in science.

To make informed decisions about which novel causal hypotheses to investigate experimentally, scientists need to know how likely the hypotheses are to be true. In probabilistic terms, this means they need to have the probabilities of the hypotheses (as output by a causal discovery algorithm) be well-calibrated. Informally, we say that probabilities are well-calibrated if events predicted to occur with probability $p$ do occur about $p$ fraction of the time, for all $p$ \cite{degroot1983comparison}. In general, it is important to use calibrated probabilities when making decisions using decision theory.

In this paper, we focus on the discovery of causal Bayesian network (CBN) structure from observational data. In particular, we focus on the discovery of the causal relationships (edge types) between pairs of measured variables. If a causal arc is novel and important, it may be worthwhile to experimentally investigate it. The extent to which it is worth doing so depends in part on how high is the calibrated probability that the causal arc is present. We introduce a method to calibrate edge type probabilities in CBNs with thousands of measured variables and arbitrarily many latent variables.

The method requires the following components: (1) a method for generating initial {\bf probability estimates} of the edge types for each pair of variables; in general those estimates need not be well-calibrated, (2) the truth status of a small unbiased sample of the causal relationships in the network, which we call a {\bf calibration training set}, and (3) a {\bf calibration method} for using the uncalibrated probability estimates and the calibration training set to generate calibrated probabilities for the large number of remaining pairs of variables. 

We use a bootstrapping method \cite{efron1994introduction} for generating probability estimates of edge types. This method resamples a dataset $n$ times with replacement and learns a model for each dataset. In particular, for each dataset we use the Really Fast Causal Inference (RFCI) algorithm \cite{colombo2012learning} to estimate the underlying generative network when allowing for the possibility of latent confounders. For any given pair of nodes $(A, B)$, the probability they have a given edge type (e.g., $A \rightarrow B$) is estimated as the fraction of that edge type for $(A, B)$ in the $n$ networks. Previously, researchers have successfully applied this approach for estimating the probabilities of edge types in Bayesian networks \cite{friedman1999data}. These bootstrap estimates are not guaranteed to represent calibrated posterior probabilities, however, even in the large sample limit of the number of bootstrap samples. A key reason is that heuristic search, while practically necessary, may get stuck in local maxima. Thus, there is a need to map those estimates to calibrated probabilities, which is the focus of the current paper.

%changed first sent by Fattaneh
The bootstrapping approach described above provides empirical estimates of edge-type posterior probabilities for both constraint-based (e.g., PC, FCI \cite{spirtes2010introduction}, and RFCI \cite{colombo2012learning}) and Bayesian structure learning algorithms (e.g., GES \cite{chickering2003optimal}). Bayesian model averaging \cite{madigan1995bayesian,friedman2003being,koivisto2004exact,eaton2007exact,koivisto2012advances} provides an alternative approach for estimating edge probabilities. However, such Bayesian methods are typically applicable when using datasets in which the number of random variables is in the double digits (for exact search methods) to triple digits (for heuristic search methods). In contrast, we are interested in providing calibrated estimates of edge probabilities for datasets that may contain thousands of variables, as typically encountered with modern biological data.  We also note that Bayesian model averaging methods are sensitive to the method applied for heuristic search \cite{friedman1999data} and to the structure and parameter priors that are used, even if they are non-informative. Consequently, their generated probabilities are still subject to possibly being uncalibrated. Finally, there are no computationally tractable Bayesian methods for discovering CBNs that contain more than a few latent confounders; in contrast, constraint-based methods exist that can perform discovery of CBNs with hundreds (or more) latent variables on datasets with thousands of variables in a feasible amount of time \cite{colombo2012learning}.

We assume the availability of a calibration training set that allows us to induce a mapping from bootstrap probability estimates to calibrated posterior probabilities. The training set should contain the truth status for the subset of edge types. In the domain of biomedical applications, the truth status might come, for example, from results published in the literature. We emphasize that the calibration training set can be very small, relative to the number of total node pairs. In the experiments we performed, it consists of less than 0.02\% of all the node pairs. Using it, our goal is to generate better calibrated probabilities for the remaining 99.98\% of node pairs. 
In an application using biomedical data, for example, a biomedical scientist who chose to experimentally test causal relationships that have high probabilities (i.e., close to 1) that are well-calibrated could be confident that the experiments would usually corroborate those relationships. We introduce a new neural-network-based calibration method that uses the calibration training set to construct a mapping from bootstrap probability estimates to calibrated posterior probabilities of edge types for all node pairs in a CBN (except those few that are used for training). We apply that mapping to all of those node pairs. 

In this paper, we use simulated data to investigate two main questions\footnote{\scriptsize{Note that it is difficult to obtain gold standards for the causal relationships among the variables in large observational datasets. As a result, the use of simulated data is important and commonly done to evaluate causal discovery methods.}}. First, how calibrated are the bootstrap-derived probabilities of edge types? Second, how calibrated are the probabilities produced by our neural-network-based calibration method? Given a finite calibration training set, the latter method is not guaranteed to always output perfectly calibrated probabilities either. \textit{Our main hypothesis in this paper is that this calibration method will output probabilities that are better calibrated than are the bootstrap probabilities, while being at least as discriminative in terms of measures such as precision, recall, and F1 score}.

\def\CircleArrow{\kern-1.5pt\hbox{$\circ$}\hbox{$\rightarrow$}}
\def\CircleCircle{\kern-1.5pt\hbox{$\circ$}\hbox{--}\hbox{$\circ$}}

\newcommand\independent{\protect\mathpalette{\protect\independenT}{\perp}}
\def\independenT#1#2{\mathrel{\rlap{$#1#2$}\mkern2mu{#1#2}}}

\section{Method}
\label{sec:method}
In this section we briefly describe the RFCI search, bootstrap RFCI, and calibration model.

\subsection{Overview of RFCI}
Colombo et al.\cite{colombo2012learning} developed an algorithm called Really Fast Causal Inference (RFCI), which identifies the causal structure of the data-generating process in the presence of latent variables using Partial Ancestral Graphs (PAGs) as a representation. A PAG encodes a Markov equivalence class of Bayesian networks (possibly with latent variables) that exhibit the same conditional independence relationships. RFCI has two stages: (1) adjacency search: this involves a selective search for the (in)dependencies among the measured variables, (2) orientation phase: this orients the endpoints among pairs of nodes that are connected according to the first stage. As is typical of constraint-based causal discovery algorithms, RFCI outputs a single graph structure (PAG) and does not provide any information about the uncertainty of the edges between the nodes in the structure.

% Colombo et al. \cite{colombo2012learning} developed an algorithm called Really Fast Causal Inference (RFCI), which identifies the causal structure of the data-generating process in the presence of latent variables using Partial Ancestral Graphs (PAGs) as a representation. A PAG represents a Markov equivalence class of DAGs (possibly with latent variables) that exhibit the same conditional independence relationships. 

% RFCI is a two-phase constraint-based algorithm that includes an adjacency search followed by an orientation phase. The first phase of RFCI starts with a fully connected undirected graph, $\mathcal{G}$. For each adjacency $A$ --- $B$ in $\mathcal{G}$, if it finds a subset of adjacent nodes $C$ that make $A$ and $B$ independent conditioned on C (i.e., $A\independent B | C$), it deletes the edge and stores $C$. In the second phase, it applies orientation rules to orient the endpoints based on the results of the first phase. As is typical of constraint-based causal discovery algorithms, RFCI outputs a single graph structure (PAG) and does not provide any information about the uncertainty of the edges between the nodes in the structure. %For more information about RFCI see \cite{colombo2012learning}.

\subsection{Bootstrap RFCI}
\label{sec:BGES}
Considering the PAG generated by RFCI, it is possible to partition all pairs of nodes $(A, B)$ into the following seven classes: (1) $A \cdots B$: there is no edge between $A$ and $B$; % The absence of an edge between $A$ and $B$ indicates that neither is a cause of the other and they are not confounded by some latent variable(s).
(2) $A\rightarrow B$: a directed edge from $A$ to $B$ means that $A$ is a direct or indirect cause of $B$;
(3) $B \rightarrow A$: this is similar to (2);
(4) $A$ $\CircleArrow$ $B$: this edge type indicates either $A$ is a cause of $B$, there is an unmeasured confounder of $A$ and $B$, or both;
(5)$B$ $\CircleArrow$ $A$: this is similar to (4);
(6) $A$ $\CircleCircle$ $B$: this edge type expresses that $A$ is a cause of $B$, $B$ is a cause of $A$, there is an unmeasured confounder of $A$ and $B$, or that there is an unmeasured confounder and one of those two causal relationships holds;
(7) $A \leftrightarrow B$: a bi-directed edge between $A$ and $B$ represents the presence of an unmeasured confounder of $A$ and $B$.

The bootstrap RFCI (BRFCI) method that we apply has three main steps. First, it performs bootstrap sampling over the training data $n$ times ($n = 200$ in our experiments) to create $n$ different bootstrap training datasets. In the second step, it runs RFCI on each of $n$ datasets to obtain $n$ PAGs. Finally, for every pair of nodes, it uses the frequency counts of each edge class for that pair over the generated PAGs to determine a probability distribution for the seven possible edge classes. As mentioned, these bootstrap estimates are not guaranteed to be calibrated. In the following section, we describe a post-processing method to map the bootstrap probabilities to calibrated probabilities.

\subsection{Calibration Model}
\label{sec:mehtod:calibration}
For a pair of nodes $(A, B)$, the resulting output of the BRFCI method will be seven jointly exhaustive and mutually exclusive class probabilities that correspond to the seven classes described above. Therefore, we need to apply a calibration method that post-processes a multi-class classification score (in our case seven classes). One simple approach to devise such a multi-class calibration model is to use a well-performing non-parametric binary classifier calibration method such as isotonic regression \cite{zadrozny2002transforming}, averaging over Bayesian binning (ABB) \cite{pakdaman20015binary}, or Bayesian binning into quantiles (BBQ) \cite{pakdaman2015obtaining} to post-process the corresponding output probabilities of each class separately. This is performed in a one-versus-remainder fashion as described in \cite{zadrozny2002transforming}. The major drawback of this approach is that such binary calibration methods are histogram-based non-parametric and they require a considerable amount of data to produce well-calibrated probabilities. However, it is often too expensive or not feasible to obtain the truth status for a large number of node pairs in real applications of causal discovery. Consequently, the availability of only a small calibration training set is a critical constraint in the design of the calibration approach.

\begin{figure}[htbp]
\begin{minipage}{.59\textwidth}
\centering
\includegraphics[width=6cm]{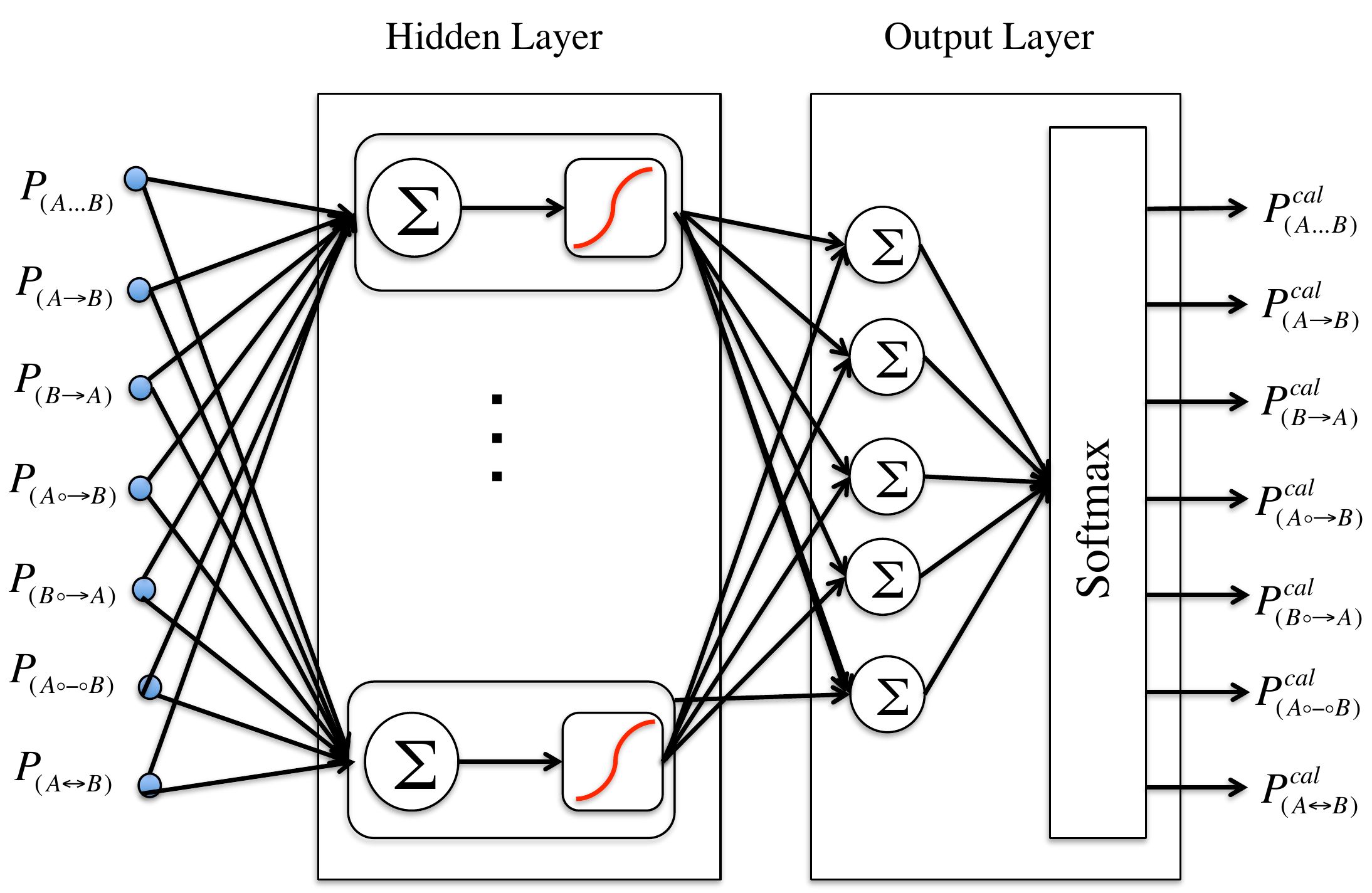}
\caption{\small{The structure of post-processing calibration method. The inputs on the left are the bootstrap probabilities for seven edge types.  The outputs on the right are the corresponding post-processed probabilities that are intended to be better calibrated.%Such an input is provided for each pair of nodes in the training set.
}} \label{fig:MLP}
\end{minipage}
\hspace{3mm}
\begin{minipage}{.37\textwidth}
\centering
\includegraphics[width=4.9cm]{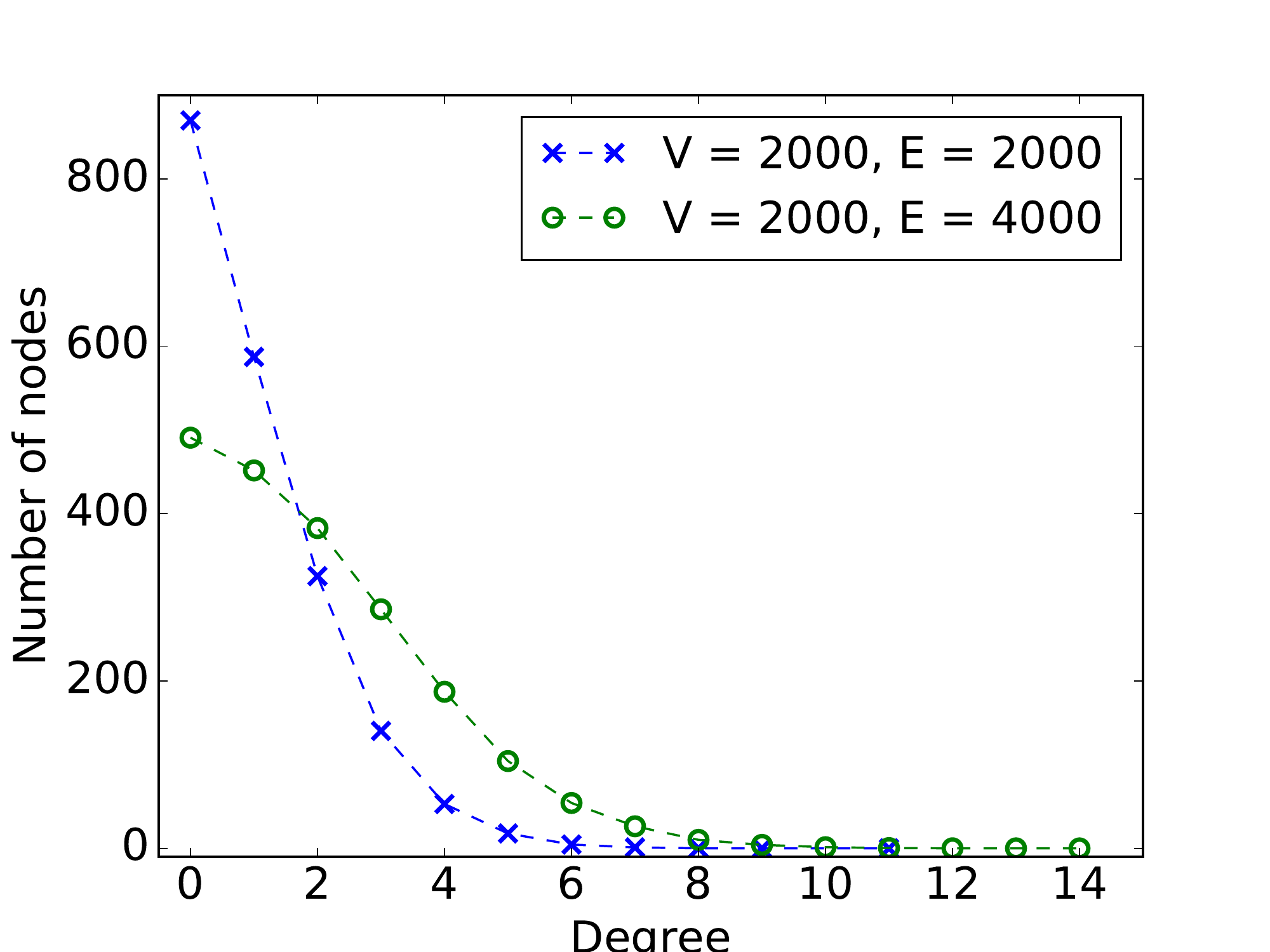}
\caption{\small{Parent size distribution for simulated CBNs with $V=2000$ nodes and $E=2000$ or 4000 edges.}}
\label{fig:CBNParentDist}
\end{minipage}
\vspace{-5mm} 
\end{figure}

To resolve this problem, we make a simple extension to Platt's method \cite{platt1999probabilistic}, which is a parametric binary classifier calibration approach. Platt's method uses a sigmoid transformation to map the output of a binary classifier into a calibrated probability. It then uses a logistic loss function to learn the two parameters of the model. The method has two advantages: (1) it has only two parameters that make it a viable choice for low sample size calibration datasets, and (2) the method runs in $O(1)$ at test time, and thus, it is fast. A natural extension to Platt's method for the multi-class calibration task is to use a combination of a softmax transfer function and a cross-entropy loss function instead of a sigmoid function and a logistic loss function, respectively. Minimizing the cross entropy is equivalent to minimizing the empirical Kullback-Leibler divergence of the estimated probabilities and the observed ones. The minimum will be achieved by the true probability distribution and minimizing the cross entropy function will result in finding the closest distribution parameterized by the model to the observed distribution of data \cite{nielsen2015neural}. 

The model that uses the softmax transfer function and optimizes the cross entropy loss function is called softmax regression \cite{nielsen2015neural}. The softmax regression-based calibration model inherits the desirable properties of Platt's method. However, similar to Platt's method, the mapping that the softmax regression-based calibration method can learn is restrictive since the final separating boundaries between each pair of classes are always linear. A simple relaxation of this restriction is to use a shallow neural network with one hidden layer. Figure \ref{fig:MLP} shows the architecture of the shallow neural network model that we use to post-process the bootstrap-generated probabilities. 

In our experiments, we train 10 different such shallow neural networks by setting the number of neurons in the hidden layer to be 4, 5, 6, or 7 randomly. At test time, we use the average of the 10 different outputs generated by these models as the final calibrated probability estimates. The averaging is helpful since it reduces the variance error of the predictions and improves the final performance of the post-processed probabilities \cite{domingos2012few}. We will use the notation $f_{cal} (p1, p2, p3, p4, p5, p6, p7)$ to denote the mapping from a seven-element vector of uncalibrated probabilities that are input to a seven-element vector of calibrated probabilities that are output. We implemented our model using the scikit flow Python package\footnote{\scriptsize{https://github.com/tensorflow/skflow}}, which uses the tensorflow machine learning package \cite{tensorflow2015-whitepaper}. We used the cross-entropy loss function and the adagrad optimization method \cite{duchi2011adaptive} to learn the parameters; we set the learning rate and the batch size to be 0.1 and 10, respectively.

\renewcommand*\floatpagefraction{0.8}
\section{Experimental Methods}
\label{sec:setup}
This section describes the experimental methods that we used to evaluate the performance of the calibrated network discovery method introduced above. The evaluation involves the following steps:

1. Create a random causal Bayesian network, $BN$, with $V$ real-valued nodes and $E$ edges. We set $V$ to be 1000 and 2000. We also set the average number of edges per node to be 1 and 2 (i.e., $E = V$ and $E=2V$). To construct the $BN$, we first ordered the nodes. Then, we randomly added edges in a forward direction until obtaining the specified mean graph density. This process generates a graph with a power-law-type distribution over the number of parents, with some nodes having many more than the average number of parents (Figure \ref{fig:CBNParentDist}). In each $BN$, the nodes correspond to continuous random variables where for every pair of nodes, $(A,B)$, we parametrize a relation $A \rightarrow B$ in the $BN$ as a structural equation model (SEM): $A = \epsilon_A$ and $B = A \beta + \epsilon_B$, where $\epsilon_A$ and $\epsilon_B$ are zero-mean Gaussian noise terms and $\beta$ is a linear coefficient. In our experiments, similar to Ramsey\cite{ramsey2015scaling,silva2006learning}, variances of $\epsilon_A$ and $\epsilon_B$ are uniformly randomly chosen from the interval $[1.0, 3.0]$ and $\beta$ is drawn uniformly randomly from the interval $[-1.5, 0.2] \cup [0.2, 1.5]$. This choice of parameter values for the simulations implies that, on average, around half of the variance of the variables is due to the error term, which makes structure learning more difficult \cite{ramsey2015scaling,silva2006learning}.

2. Simulate a dataset $D$ of size 1000 from $BN$, subject to constraints that are described below. 

3. Set a percentage of variables, $h$, to be unobserved (i.e., latent). These latent variables are randomly chosen from confounder variables (i.e., common causes) in a given data-generating $BN$. We set $h$ to be either 10\% or 20\%.

4. Generate 200 bootstrap datasets, $DB[1..200]$ from $D$. 

5. For each bootstrap dataset, learn a PAG using the RFCI method; let $PAG[1..200]$ designate these PAGs. RFCI uses Fisher's Z test to check conditional independence of variables in the dataset. We set the significance level at which independence judgments were made to be $\alpha = 0.001$ and $0.005$. 

6. For each node pair $(A, B)$, calculate the probability distribution $P_e(A, B)$ of the edge types of $(A, B)$ using maximum likelihood estimates on the counts in $PAG[1..200]$.

7. Perform a stratified random sampling on the node pairs to obtain $N$ training samples for calibration and use the rest of the data for testing. We set $N= 70, 140,\text{ or } 210$. In obtaining $N$ samples, we used stratified random sampling to select $N$/7 samples for each of the seven edge classes\footnote{\scriptsize{Using stratified random sampling is crucial due to severe class imbalance of the data (i.e., more than 99\% of the pairs are no-edge type).}}. In particular, we first sorted the probability scores of edges in each edge class according to the bootstrap probabilities. We then partitioned the instances into $5$ bins of $\{[0,0.2),[0.2,0.4),[0.4,0.6),[0.6,0.8),[0.8,1]\}$ based on their bootstrap probabilities. Finally, we sampled separately from each bin with equal frequency.

8. Learn the calibration function $f_{cal}$ using the calibration training data.

9. For each node pair $(A, B)$ in the test set, derive $P_e^{cal}(A,B) = f_{cal}\left(P_e(A, B)\right)$.

10. Compare the performance of $P_e^{cal}$ versus $P_e$ in correctly predicting the data-generating structure of $BN$ for the test set pairs, and doing so in a manner that is well calibrated.

In running the above evaluation procedure, step (5) is the most time consuming part that involves running RFCI on 200 bootstrap datasets. However, this is still feasible due to the run-time efficiency of the RFCI method and our use of parallel computing\footnote{\scriptsize{The running times of the experiments varied from 11 to 176 minutes on a 16-core compute node, which is computationally feasible, because step (5) needs to be done one time only.}}. For all simulations, we used Tetrad\footnote{\scriptsize{https://github.com/cmu-phil/tetrad}}, which is an open-source, freely available software application that is coded in Java.
%Greg: what is "it"?

Steps (1) through (10) above were repeated for 10 randomly generated BNs and the performance results were averaged. For a given node pair, we take the predicted edge type for that pair to be the one that has the highest probability. Note that although there are seven different edge classes, we consider only five edge types for performance evaluation, because $A \rightarrow B$ and $B \rightarrow A$ are both directed edge types, and $A$ $\CircleArrow$ $B$ and $B$ $\CircleArrow$ $A$ are both partially directed edge types.

The first two edge-type-based evaluation measures are \textit{precision} (P) and \textit{recall} (R). To compute these measures for each edge type, we calculated the four basic statistics of true positives (TP), false positives (FP), true negatives (TN), and false negatives (FN) for each of the types separately. Precision is then derived as the ratio $TP/(TP+FP)$. Recall is derived as the ratio $TP/(TP+FN)$. We also report the F1-score (i.e., the harmonic mean of the precision and recall), which is a summary measure that shows the overall performance of the predictions in terms of both precision and recall.

% Greg: how 100 instances are chosen?
% Mahdi: We assume 100 instance is large enough to trust the counts inside each bin!
We also evaluated the edge-type-based predictions in terms of maximum calibration error (MCE) \cite{pakdaman2015obtaining}. We calculated the MCE for each edge type by partitioning the output space of the estimated edge-type probabilities, which is on the interval [0, 1], into equal-frequency bins with 100 randomly chosen instances. The estimated probability for each instance is located in one of the bins. For each bin, we define the associated calibration error as the absolute difference between the mean value of the predictions and the actual observed frequency of positive instances. The MCE calculates the maximum calibration error over all the bins. The lower the value of MCE, the better the calibration of the probability scores. The lowest possible value of MCE is 0 and the highest possible value is 1.

%Greg: Not very clear. Can this text be clarified, simplified, and shortened?
%We just wanted to have a single number for comparison for the seven classes and a micro average MCE is one reasonable summary.(refer to macro vs micro average F1 in multi label classification)
We also report the overall (micro averaged) MCE as a summary measure which shows the performance of the predictions in terms of calibration. To compute this measure, we augmented the seven-element probability distribution vectors, $P_{e}(A,B)$ for all test instances to form an aggregated vector $P_{all}$. We also augmented their corresponding 1-of-k binary labels \cite{nielsen2015neural,tensorflow2015-whitepaper} to form an aggregated binary vector $Z_{all}$. The overall MCE is defined as the maximum calibration error calculated based on $P_{all}$ and $Z_{all}$.

\section{Experimental Results}
\label{sec:results}
\vspace{-3.1mm}
This section presents the results of our experiments in evaluating the performance of the generated probabilities for the five edge types before and after calibration. We use the shallow neural network calibration method to learn the calibration function $f_{cal}$ from calibration training data. Since the purpose of this paper is not to compare calibration methods, we do not report the results of experiments on using other calibration methods (e.g., IsoReg or Platt's method). Rather, we only report the results of calibration using the neural network method which we found performs well with relatively small calibration training sample sizes compared to the other calibration methods that we tried.

For each set of configurations (e.g., $N=210$, $V = 2000$, $E = 2000$), we report the average results of using 10 randomly simulated CBNs. Tables \ref{tab:v2K_h0.1_alpha0.001}, \ref{tab:v2K_h0.2_alpha0.001}, \ref{tab:v2K_h0.1_alpha0.005}, and \ref{tab:v2K_h0.2_alpha0.005} show the results for CBNs with 2000 nodes (due to the page limit, the results of experiments for $V = 1000$ are not included but similar results are achieved). In these tables, boldface indicates the results that are statistically significantly superior, based on a two-sided Wilcoxon signed rank test at 5\% significance level. Tables \ref{tab:v2K_h0.1_alpha0.001}-\ref{tab:v2K_h0.2_alpha0.005} indicate that by post-processing the bootstrap probabilities, we can improve the overall edge-type performance both in terms of discrimination and calibration. The only exception is the $A \leftrightarrow B$ edge type for which we lose discrimination. This is happening because the original precision and recall of the bootstrap probabilities are very low for this edge type. Consequently, we often obtain very few positive instances from this edge type in the calibration training set, which negatively affects the performance of predictions after calibration. Note that for the no-edge type we do not report precision, recall, and F1, because they were always very close to 1.
%Greg: Do you mean predicted instances or true (gold standard) instances? Please clarify.
\begin{table*}[htbp]
\vspace{-10mm}
\small
\centering
\caption{\small{Simulation results. $V$, $E$, and $h$ represent the number of variables, edges, and percentage of hidden variables in the data-generating CBN, respectively. $\alpha$ is the significance level used in the RFCI method and $N$ is the calibration training set size. Boldface indicates the results that are significantly better, based on the Wilcoxon signed rank test at 5\% significance level. For MCE, lower is better.}} \label{tab:results}
\scalebox{0.67}{
\subfloat[$V = 2000$, $h = 0.1$, and $\alpha = 0.001$]
{
	\begin{tabular}
{K{0.4cm}|c|l|K{0.3cm}K{0.3cm}K{0.3cm}K{0.6cm}|K{0.3cm}K{0.3cm}K{0.3cm}K{0.6cm}|K{0.3cm}K{0.3cm}K{0.3cm}K{0.6cm}|K{0.3cm}K{0.3cm}K{0.3cm}K{0.6cm}|c|K{0.65cm}}
    \toprule
    \multirow{2}[0]{*}{$N$} & \multirow{2}[0]{*}{$E$} & \multirow{2}[0]{*}{method} & \multicolumn{4}{c|}{\textit{A} $\rightarrow$ \textit{B}}    & \multicolumn{4}{c|}{\textit{A} $\CircleArrow$ \textit{B}}   & \multicolumn{4}{c|}{\textit{A} $\CircleCircle$ \textit{B}}   & \multicolumn{4}{c|}{\textit{A} $\leftrightarrow$ \textit{B}}   & \textit{A} $\cdots$ \textit{B}   & Overall \\
    \cline{4-20}
          &       &       & P & R & F1    & MCE   & P & R & F1    & MCE   & P & R & F1    & MCE   & P & R & F1    & MCE   & MCE   &  MCE \\
    \hline
    \hline
    \multirow{4}[0]{*}{70} & \multirow{2}[0]{*}{2K} & before & 0.61  & 0.33  & 0.42  & 0.12  & 0.45  & 0.07  & 0.12  & 0.44  & 0.79  & 0.06  & 0.12  & 0.65  & 0.05  & \textbf{0.25} & \textbf{0.08} & 0.94  & 0.22  & 0.33 \\
          &       & after & \textbf{0.69} & \textbf{0.37} & \textbf{0.47} & 0.10  & \textbf{0.57} & \textbf{0.62} & \textbf{0.59} & \textbf{0.25} & 0.79  & \textbf{0.44} & \textbf{0.56} & \textbf{0.27} & 0.03  & 0.02  & 0.02  & \textbf{0.18} & \textbf{0.14} & \textbf{0.23} \\
     \cline{2-21}
          & \multirow{2}[0]{*}{4K} & before & 0.66  & 0.30  & 0.41  & 0.30  & \textbf{0.51} & 0.03  & 0.06  & 0.45  & 0.62  & 0.03  & 0.05  & 0.44  & \textbf{0.04} & \textbf{0.09} & \textbf{0.06} & 0.95  & 0.51  & 0.55 \\
          &       & after & 0.67  & \textbf{0.38} & \textbf{0.48} & \textbf{0.20} & 0.46  & \textbf{0.43} & \textbf{0.44} & \textbf{0.23} & 0.68  & \textbf{0.27} & \textbf{0.38} & \textbf{0.26} & 0.00  & 0.00  & 0.00  & \textbf{0.09} & \textbf{0.21} & \textbf{0.22} \\
    \cline{1-21}
    \multirow{4}[0]{*}{140} & \multirow{2}[0]{*}{2K} & before & 0.57  & 0.28  & 0.37  & 0.10  & 0.42  & 0.06  & 0.10  & 0.46  & 0.64  & 0.05  & 0.09  & 0.65  & 0.04  & \textbf{0.23} & \textbf{0.07} & 0.93  & 0.21  & 0.34 \\
          &       & after & \textbf{0.66} & \textbf{0.31} & \textbf{0.41} & 0.09  & \textbf{0.58} & \textbf{0.63} & \textbf{0.60} & \textbf{0.27} & 0.79  & \textbf{0.41} & \textbf{0.54} & \textbf{0.29} & 0.02  & 0.01  & 0.01  & \textbf{0.17} & \textbf{0.12} & \textbf{0.26} \\
    \cline{2-21}
          & \multirow{2}[0]{*}{4K} & before & 0.66  & 0.29  & 0.40  & 0.30  & 0.51  & 0.03  & 0.06  & 0.45  & 0.44  & 0.02  & 0.03  & 0.44  & \textbf{0.04} & \textbf{0.09} & \textbf{0.06} & 0.95  & 0.51  & 0.55 \\
          &       & after & \textbf{0.67} & \textbf{0.37} & \textbf{0.48} & \textbf{0.17} & 0.47  & \textbf{0.44} & \textbf{0.46} & \textbf{0.24} & \textbf{0.68} & \textbf{0.26} & \textbf{0.37} & \textbf{0.26} & 0.00  & 0.00  & 0.00  & \textbf{0.09} & \textbf{0.18} & \textbf{0.21} \\
    \cline{1-21}
    \multirow{4}[0]{*}{210} & \multirow{2}[0]{*}{2K} & before & 0.53  & 0.24  & 0.32  & 0.10  & 0.40  & 0.05  & 0.09  & 0.46  & 0.62  & 0.04  & 0.07  & 0.66  & 0.04  & \textbf{0.23} & \textbf{0.07} & 0.93  & 0.21  & 0.35 \\
          &       & after & \textbf{0.65} & \textbf{0.28} & \textbf{0.37} & 0.09  & \textbf{0.58} & \textbf{0.64} & \textbf{0.61} & \textbf{0.26} & 0.80  & \textbf{0.40} & \textbf{0.53} & \textbf{0.25} & 0.02  & 0.01  & 0.01  & \textbf{0.16} & \textbf{0.12} & \textbf{0.26} \\
    \cline{2-21}
          & \multirow{2}[0]{*}{4K} & before & 0.65  & 0.29  & 0.40  & 0.31  & 0.51  & 0.03  & 0.05  & 0.46  & 0.27  & 0.01  & 0.02  & 0.44  & \textbf{0.04} & \textbf{0.09} & \textbf{0.06} & 0.95  & 0.51  & 0.56 \\
          &       & after & 0.67  & \textbf{0.36} & \textbf{0.47} & \textbf{0.19} & 0.47  & \textbf{0.43} & \textbf{0.45} & \textbf{0.23} & \textbf{0.66} & \textbf{0.23} & \textbf{0.34} & \textbf{0.29} & 0.01  & 0.00  & 0.00  & \textbf{0.08} & \textbf{0.17} & \textbf{0.20} \\
    \bottomrule
    \end{tabular}%   
    \label{tab:v2K_h0.1_alpha0.001}
    }
}\\
\scalebox{0.67}{
\subfloat[$V = 2000$, $h = 0.2$, and $\alpha = 0.001$]
{
	\begin{tabular}{K{0.4cm}|c|l|K{0.3cm}K{0.3cm}K{0.3cm}K{0.6cm}|K{0.3cm}K{0.3cm}K{0.3cm}K{0.6cm}|K{0.3cm}K{0.3cm}K{0.3cm}K{0.6cm}|K{0.3cm}K{0.3cm}K{0.3cm}K{0.6cm}|c|K{0.65cm}}
    \toprule
    \multirow{2}[0]{*}{$N$} & \multirow{2}[0]{*}{$E$} & \multirow{2}[0]{*}{method} & \multicolumn{4}{c|}{\textit{A} $\rightarrow$ \textit{B}}    & \multicolumn{4}{c|}{\textit{A} $\CircleArrow$ \textit{B}}   & \multicolumn{4}{c|}{\textit{A} $\CircleCircle$ \textit{B}}   & \multicolumn{4}{c|}{\textit{A} $\leftrightarrow$ \textit{B}}   & \textit{A} $\cdots$ \textit{B}   & Overall \\
    \cline{4-20}
          &       &       & P & R & F1    & MCE   & P & R & F1    & MCE   & P & R & F1    & MCE   & P & R & F1    & MCE   & MCE   &  MCE \\
    \hline
    \hline
    \multirow{4}[0]{*}{70} & \multirow{2}[0]{*}{2K} & before & 0.35  & 0.15  & 0.21  & 0.10  & 0.37  & 0.03  & 0.05  & 0.48  & 0.56  & 0.03  & 0.05  & 0.66  & 0.11  & \textbf{0.18} & 0.13  & 0.90  & 0.24  & 0.34 \\
          &       & after & \textbf{0.43} & 0.15  & 0.21  & 0.08  & \textbf{0.53} & \textbf{0.34} & \textbf{0.41} & \textbf{0.26} & 0.77  & \textbf{0.30} & \textbf{0.42} & \textbf{0.27} & \textbf{0.21} & 0.10  & 0.12  & \textbf{0.29} & \textbf{0.13} & \textbf{0.24} \\
    \cline{2-21}
          & \multirow{2}[0]{*}{4K} & before & 0.65  & 0.25  & 0.36  & 0.28  & 0.49  & 0.02  & 0.04  & 0.44  & 0.36  & 0.02  & 0.04  & 0.40  & 0.09  & \textbf{0.06} & \textbf{0.07} & 0.91  & 0.53  & 0.56 \\
          &       & after & 0.63  & \textbf{0.33} & \textbf{0.42} & \textbf{0.18} & 0.44  & \textbf{0.31} & \textbf{0.36} & \textbf{0.25} & \textbf{0.65} & \textbf{0.21} & \textbf{0.31} & \textbf{0.22} & 0.08  & 0.01  & 0.01  & \textbf{0.14} & \textbf{0.20} & \textbf{0.23} \\
    \cline{1-21}
    \multirow{4}[0]{*}{140} & \multirow{2}[0]{*}{2K} & before & 0.32  & 0.11  & 0.15  & 0.09  & 0.32  & 0.02  & 0.04  & 0.50  & 0.44  & 0.02  & 0.03  & 0.67  & 0.10  & \textbf{0.17} & 0.13  & 0.90  & 0.23  & 0.33 \\
          &       & after & 0.39  & 0.10  & 0.14  & 0.08  & \textbf{0.55} & \textbf{0.34} & \textbf{0.41} & \textbf{0.26} & 0.79  & \textbf{0.28} & \textbf{0.40} & \textbf{0.30} & \textbf{0.20} & 0.10  & 0.11  & \textbf{0.27} & \textbf{0.12} & \textbf{0.23} \\
    \cline{2-21}
          & \multirow{2}[0]{*}{4K} & before & 0.65  & 0.24  & 0.35  & 0.28  & 0.48  & 0.02  & 0.03  & 0.45  & 0.20  & 0.01  & 0.03  & 0.40  & 0.09  & \textbf{0.06} & \textbf{0.07} & 0.91  & 0.53  & 0.59 \\
          &       & after & 0.65  & \textbf{0.31} & \textbf{0.42} & \textbf{0.15} & 0.43  & \textbf{0.31} & \textbf{0.36} & \textbf{0.24} & \textbf{0.65} & \textbf{0.22} & \textbf{0.32} & \textbf{0.20} & 0.06  & 0.01  & 0.01  & \textbf{0.14} & \textbf{0.19} & \textbf{0.21} \\
    \cline{1-21}
    \multirow{4}[0]{*}{210} & \multirow{2}[0]{*}{2K} & before & 0.26  & 0.09  & 0.13  & 0.10  & 0.32  & 0.02  & 0.03  & 0.50  & 0.40  & 0.01  & 0.02  & 0.67  & 0.10  & \textbf{0.17} & 0.13  & 0.90  & 0.23  & 0.33 \\
          &       & after & \textbf{0.35} & 0.08  & 0.13  & \textbf{0.08} & \textbf{0.55} & \textbf{0.32} & \textbf{0.40} & \textbf{0.27} & \textbf{0.80} & \textbf{0.27} & \textbf{0.39} & \textbf{0.31} & \textbf{0.19} & 0.10  & 0.11  & \textbf{0.28} & \textbf{0.12} & \textbf{0.23} \\
    \cline{2-21}
          & \multirow{2}[0]{*}{4K} & before & 0.65  & 0.23  & 0.34  & 0.29  & 0.49  & 0.01  & 0.03  & 0.45  & 0.13  & 0.01  & 0.01  & 0.40  & 0.09  & \textbf{0.06} & \textbf{0.07} & 0.91  & 0.52  & 0.58 \\
          &       & after & 0.65  & \textbf{0.30} & \textbf{0.41} & \textbf{0.15} & 0.43  & \textbf{0.31} & \textbf{0.36} & \textbf{0.23} & \textbf{0.63} & \textbf{0.20} & \textbf{0.30} & \textbf{0.20} & 0.14  & 0.01  & 0.01  & \textbf{0.12} & \textbf{0.18} & \textbf{0.21} \\
    \bottomrule
    \end{tabular}%   
    \label{tab:v2K_h0.2_alpha0.001}
    }
}\\
\scalebox{0.67}{
\subfloat[$V = 2000$, $h = 0.1$, and $\alpha = 0.005$]
{
	 \begin{tabular}{K{0.4cm}|c|l|K{0.3cm}K{0.3cm}K{0.3cm}K{0.6cm}|K{0.3cm}K{0.3cm}K{0.3cm}K{0.6cm}|K{0.3cm}K{0.3cm}K{0.3cm}K{0.6cm}|K{0.3cm}K{0.3cm}K{0.3cm}K{0.6cm}|c|K{0.65cm}}
    \toprule
    \multirow{2}[0]{*}{$N$} & \multirow{2}[0]{*}{$E$} & \multirow{2}[0]{*}{method} & \multicolumn{4}{c|}{\textit{A} $\rightarrow$ \textit{B}}    & \multicolumn{4}{c|}{\textit{A} $\CircleArrow$ \textit{B}}   & \multicolumn{4}{c|}{\textit{A} $\CircleCircle$ \textit{B}}   & \multicolumn{4}{c|}{\textit{A} $\leftrightarrow$ \textit{B}}   & \textit{A} $\cdots$ \textit{B}   & Overall \\
    \cline{4-20}
          &       &       & P     & R     & F1    & MCE   & P     & R     & F1    & MCE   & P     & R     & F1    & MCE   & P     & R     & F1    & MCE   & MCE   & MCE \\
    \hline
    \hline
    \multirow{4}[0]{*}{70} & \multirow{2}[0]{*}{2K} & before & 0.60  & 0.19  & 0.28  & 0.16  & 0.36  & 0.01  & 0.02  & 0.43  & 0.17  & 0.00  & 0.00  & 0.70  & \textbf{0.05} & \textbf{0.30} & \textbf{0.09} & 0.96  & 0.24  & 0.42 \\
          &       & after & 0.65  & \textbf{0.32} & \textbf{0.41} & \textbf{0.12} & 0.43  & \textbf{0.62} & \textbf{0.51} & \textbf{0.27} & \textbf{0.76} & \textbf{0.18} & \textbf{0.28} & \textbf{0.25} & 0.00  & 0.00  & 0.00  & \textbf{0.09} & \textbf{0.14} & \textbf{0.25} \\
   \cline{2-21}
          & \multirow{2}[0]{*}{4K} & before & \textbf{0.74} & 0.29  & 0.41  & 0.29  & 0.39  & 0.01  & 0.01  & 0.46  & 0.18  & 0.00  & 0.00  & 0.46  & 0.05  & \textbf{0.12} & \textbf{0.07} & 0.97  & 0.38  & 0.55 \\
          &       & after & 0.70  & \textbf{0.39} & \textbf{0.50} & \textbf{0.20} & 0.42  & \textbf{0.44} & \textbf{0.43} & \textbf{0.27} & \textbf{0.66} & \textbf{0.14} & \textbf{0.22} & \textbf{0.28} & 0.01  & 0.00  & 0.00  & \textbf{0.10} & \textbf{0.17} & \textbf{0.24} \\
   \cline{1-21}
    \multirow{4}[0]{*}{140} & \multirow{2}[0]{*}{2K} & before & 0.49  & 0.13  & 0.20  & 0.16  & 0.27  & 0.01  & 0.01  & 0.43  & 0.00  & 0.00  & 0.00  & 0.69  & \textbf{0.05} & \textbf{0.29} & \textbf{0.09} & 0.96  & 0.24  & 0.38 \\
          &       & after & \textbf{0.62} & \textbf{0.23} & \textbf{0.32} & \textbf{0.11} & \textbf{0.44} & \textbf{0.62} & \textbf{0.51} & \textbf{0.26} & \textbf{0.77} & \textbf{0.18} & \textbf{0.28} & \textbf{0.24} & 0.00  & 0.00  & 0.00  & \textbf{0.09} & \textbf{0.13} & \textbf{0.24} \\
     \cline{2-21}
          & \multirow{2}[0]{*}{4K} & before & \textbf{0.74} & 0.28  & 0.41  & 0.29  & 0.39  & 0.01  & 0.01  & 0.46  & 0.12  & 0.00  & 0.00  & 0.46  & \textbf{0.05} & \textbf{0.11} & \textbf{0.06} & 0.97  & 0.38  & 0.55 \\
          &       & after & 0.70  & \textbf{0.40} & \textbf{0.50} & \textbf{0.19} & 0.44  & \textbf{0.44} & \textbf{0.44} & \textbf{0.27} & \textbf{0.73} & \textbf{0.16} & \textbf{0.26} & \textbf{0.27} & 0.00  & 0.00  & 0.00  & \textbf{0.09} & \textbf{0.17} & \textbf{0.25} \\
     \cline{1-21}
    \multirow{4}[0]{*}{210} & \multirow{2}[0]{*}{2K} & before & 0.33  & 0.07  & 0.11  & 0.16  & 0.15  & 0.00  & 0.00  & 0.44  & 0.00  & 0.00  & 0.00  & 0.62  & \textbf{0.05} & \textbf{0.29} & \textbf{0.09} & 0.96  & 0.24  & 0.34 \\
          &       & after & 0.43  & \textbf{0.14} & \textbf{0.19} & \textbf{0.11} & \textbf{0.46} & \textbf{0.63} & \textbf{0.53} & \textbf{0.26} & \textbf{0.76} & \textbf{0.15} & \textbf{0.24} & \textbf{0.21} & 0.00  & 0.00  & 0.00  & \textbf{0.09} & \textbf{0.11} & \textbf{0.25} \\
     \cline{2-21}
          & \multirow{2}[0]{*}{4K} & before & \textbf{0.75} & 0.27  & 0.40  & 0.30  & 0.35  & 0.00  & 0.01  & 0.47  & 0.10  & 0.00  & 0.00  & 0.47  & \textbf{0.05} & \textbf{0.11} & \textbf{0.06} & 0.97  & 0.38  & 0.54 \\
          &       & after & 0.72  & \textbf{0.37} & \textbf{0.48} & \textbf{0.18} & 0.44  & \textbf{0.45} & \textbf{0.44} & \textbf{0.26} & \textbf{0.73} & \textbf{0.17} & \textbf{0.27} & \textbf{0.25} & 0.00  & 0.00  & 0.00  & \textbf{0.08} & \textbf{0.15} & \textbf{0.24} \\
    \bottomrule
    \end{tabular}%
    \label{tab:v2K_h0.1_alpha0.005}
    }
}\\
\scalebox{0.67}{
\subfloat[$V = 2000$, $h = 0.2$, and $\alpha = 0.005$]
{
	  \begin{tabular}{K{0.4cm}|c|l|K{0.3cm}K{0.3cm}K{0.3cm}K{0.6cm}|K{0.3cm}K{0.3cm}K{0.3cm}K{0.6cm}|K{0.3cm}K{0.3cm}K{0.3cm}K{0.6cm}|K{0.3cm}K{0.3cm}K{0.3cm}K{0.6cm}|c|K{0.68cm}}
    \toprule
    \multirow{2}[0]{*}{$N$} & \multirow{2}[0]{*}{$E$} & \multirow{2}[0]{*}{method} & \multicolumn{4}{c|}{\textit{A} $\rightarrow$ \textit{B}}    & \multicolumn{4}{c|}{\textit{A} $\CircleArrow$ \textit{B}}   & \multicolumn{4}{c|}{\textit{A} $\CircleCircle$ \textit{B}}   & \multicolumn{4}{c|}{\textit{A} $\leftrightarrow$ \textit{B}}   & \textit{A} $\cdots$ \textit{B}   & Overall \\
    \cline{4-20}
          &       &       & P     & R     & F1    & MCE   & P     & R     & F1    & MCE   & P     & R     & F1    & MCE   & P     & R     & F1    & MCE   & MCE   &  MCE \\
    \hline
    \hline
    \multirow{4}[0]{*}{70} & \multirow{2}[0]{*}{2K} & before & 0.25  & 0.05  & 0.07  & 0.12  & 0.17  & 0.00  & 0.01  & 0.42  & 0.06  & 0.00  & 0.00  & 0.49  & 0.11  & \textbf{0.21} & \textbf{0.15} & 0.92  & 0.22  & 0.52 \\
          &       & after & 0.39  & \textbf{0.10} & \textbf{0.14} & 0.09  & \textbf{0.39} & \textbf{0.45} & \textbf{0.41} & \textbf{0.24} & \textbf{0.67} & \textbf{0.12} & \textbf{0.19} & \textbf{0.20} & 0.06  & 0.01  & 0.02  & \textbf{0.14} & \textbf{0.13} & \textbf{0.23} \\
    \cline{2-21}
          & \multirow{2}[0]{*}{4K} & before & \textbf{0.72} & 0.22  & 0.34  & 0.30  & 0.32  & 0.00  & 0.01  & 0.45  & 0.12  & 0.00  & 0.00  & 0.43  & 0.10  & \textbf{0.08} & \textbf{0.09} & 0.94  & 0.40  & 0.49 \\
          &       & after & 0.68  & \textbf{0.30} & \textbf{0.42} & \textbf{0.22} & 0.37  & \textbf{0.32} & \textbf{0.35} & \textbf{0.27} & \textbf{0.70} & \textbf{0.12} & \textbf{0.20} & \textbf{0.25} & \textbf{0.05} & 0.00  & 0.01  & \textbf{0.13} & \textbf{0.18} & \textbf{0.25} \\
    \cline{1-21}
    \multirow{4}[0]{*}{140} & \multirow{2}[0]{*}{2K} & before & 0.15  & 0.02  & 0.03  & 0.11  & 0.14  & 0.00  & 0.00  & 0.44  & 0.00  & 0.00  & 0.00  & 0.47  & 0.11  & \textbf{0.21} & \textbf{0.14} & 0.92  & 0.22  & 0.50 \\
          &       & after & 0.24  & \textbf{0.06} & \textbf{0.09} & \textbf{0.07} & \textbf{0.40} & \textbf{0.45} & \textbf{0.42} & \textbf{0.25} & \textbf{0.72} & \textbf{0.11} & \textbf{0.17} & \textbf{0.20} & 0.09  & 0.01  & 0.02  & \textbf{0.14} & \textbf{0.12} & \textbf{0.23} \\
    \cline{2-21}
          & \multirow{2}[0]{*}{4K} & before & \textbf{0.73} & 0.21  & 0.32  & 0.31  & 0.26  & 0.00  & 0.00  & 0.46  & 0.00  & 0.00  & 0.00  & 0.39  & \textbf{0.10} & \textbf{0.08} & \textbf{0.08} & 0.94  & 0.39  & 0.48 \\
          &       & after & 0.71  & \textbf{0.29} & \textbf{0.41} & \textbf{0.20} & 0.38  & \textbf{0.33} & \textbf{0.35} & \textbf{0.25} & \textbf{0.67} & \textbf{0.12} & \textbf{0.20} & \textbf{0.21} & 0.04  & 0.00  & 0.01  & \textbf{0.12} & \textbf{0.15} & \textbf{0.23} \\
    \cline{1-21}
    \multirow{4}[0]{*}{210} & \multirow{2}[0]{*}{2K} & before & 0.07  & 0.00  & 0.01  & 0.11  & 0.06  & 0.00  & 0.00  & 0.47  & 0.00  & 0.00  & 0.00  & 0.39  & 0.12  & \textbf{0.20} & \textbf{0.15} & 0.92  & 0.22  & 0.48 \\
          &       & after & 0.13  & 0.02  & 0.03  & \textbf{0.07} & \textbf{0.40} & \textbf{0.45} & \textbf{0.42} & \textbf{0.24} & \textbf{0.66} & \textbf{0.06} & \textbf{0.10} & \textbf{0.20} & 0.09  & 0.01  & 0.02  & \textbf{0.13} & \textbf{0.11} & \textbf{0.23} \\
    \cline{2-21}
          & \multirow{2}[0]{*}{4K} & before & \textbf{0.73} & 0.20  & 0.31  & 0.32  & 0.26  & 0.00  & 0.00  & 0.46  & 0.00  & 0.00  & 0.00  & 0.37  & \textbf{0.10} & \textbf{0.08} & \textbf{0.08} & 0.94  & 0.40  & 0.48 \\
          &       & after & 0.71  & \textbf{0.28} & \textbf{0.40} & \textbf{0.21} & 0.38  & \textbf{0.33} & \textbf{0.35} & \textbf{0.26} & \textbf{0.71} & \textbf{0.12} & \textbf{0.20} & \textbf{0.21} & 0.03  & 0.00  & 0.00  & \textbf{0.12} & \textbf{0.16} & \textbf{0.25} \\
    \bottomrule
    \end{tabular}%
    \label{tab:v2K_h0.2_alpha0.005}
    }
}
% \vspace{-23mm}
\end{table*}

Figure \ref{fig:calibration_curve} shows the calibration diagram \cite{degroot1983comparison} of the estimated probabilities before and after calibration when we use 210 calibration training instances. We emphasize that observing 210 calibration instances is equivalent to observing less than 0.02\% of all node pairs in the CBN (i.e., there are $1999 \times 10^3$ node pairs in a CBN with 2000 nodes). To draw the calibration diagrams, we partitioned the output space of the estimated probabilities into five equal-size bins. In each bin, we draw the average frequency of positive class versus the mean of the predictions that are located in that bin. In these diagrams, the straight dashed line connecting (0, 0) to (1, 1) represents a perfectly calibrated model. The closer a calibration curve is to this line, the better calibrated is the prediction model.
\begin{table}[htbp]
\vspace{-22mm}
  \centering
  \scalebox{0.75}{
  \begin{tabular}
 {K{1cm}K{3cm}K{3cm}K{3cm}K{3cm}K{3cm}}
  
   & $A \rightarrow B$ & $A \CircleArrow B$  & $A \CircleCircle B$  & $A \leftrightarrow B$ & $A \cdots B$
  \\
  \small{V = 2k, E = 2k}
     & 
    \begin{minipage}{.2\textwidth}
    \includegraphics[width=2.5cm,height=2.5cm]{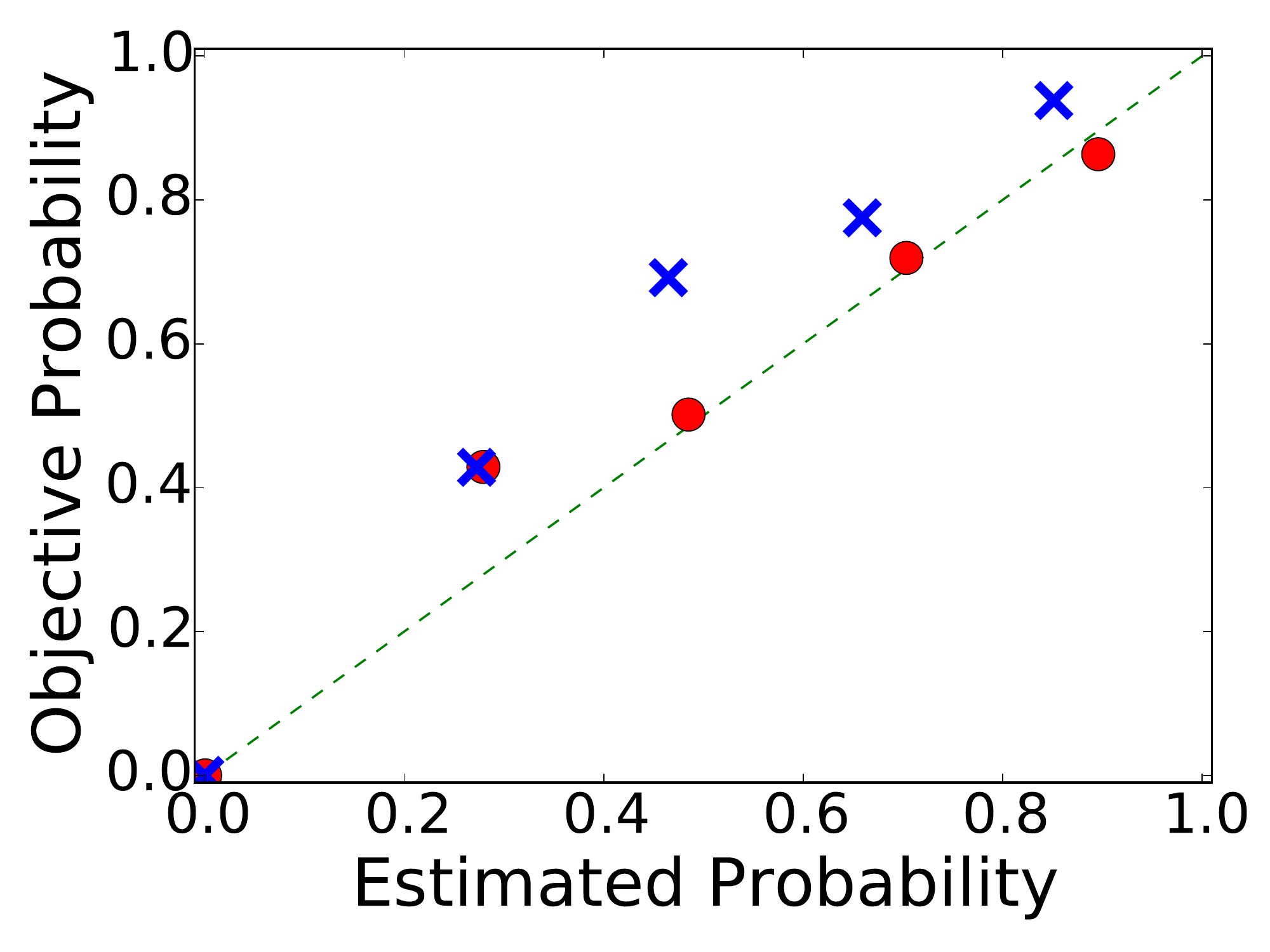}
    \end{minipage}
     & 
     \begin{minipage}{.2\textwidth}
      \includegraphics[width=2.5cm,height=2.5cm]{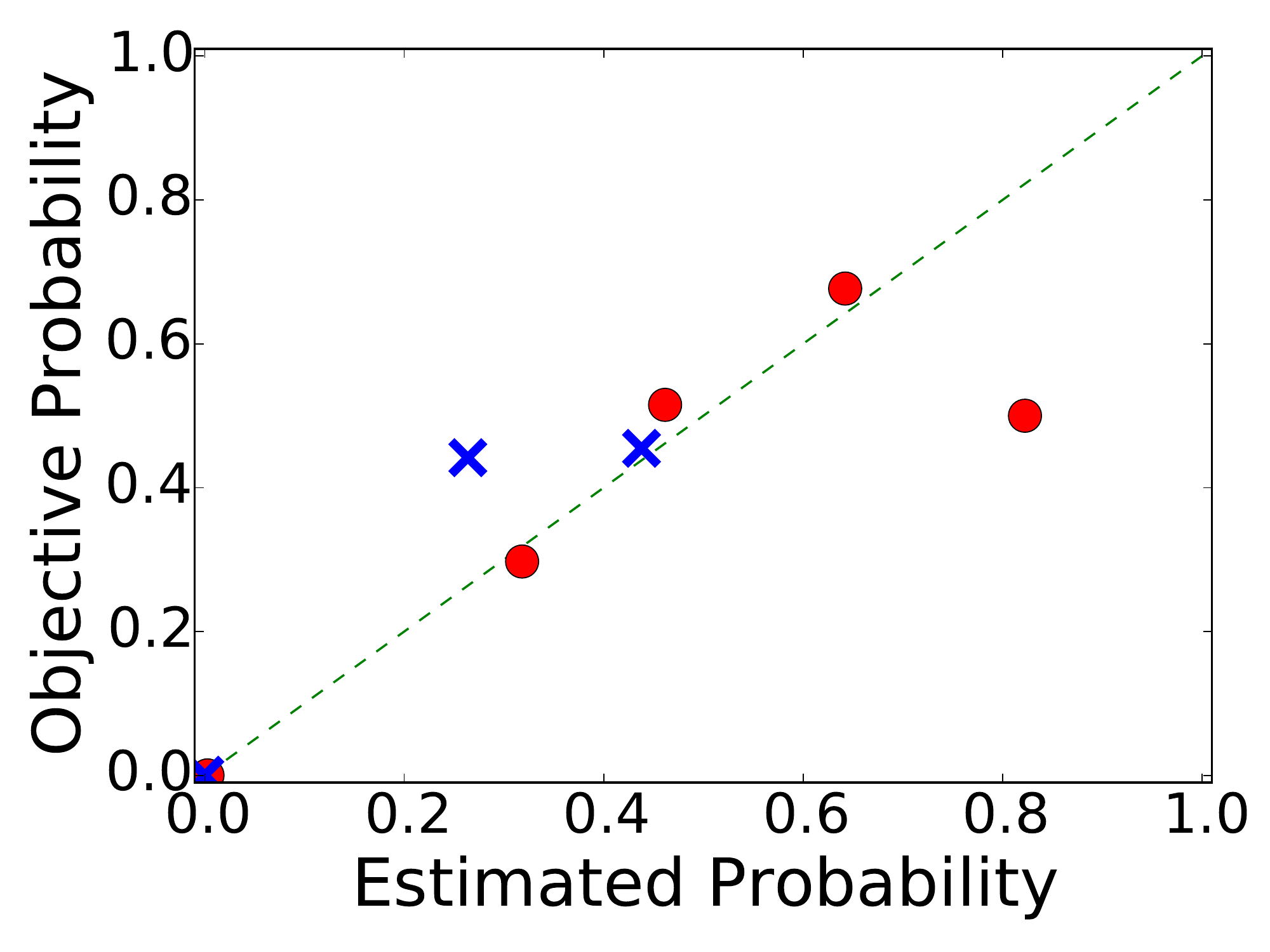}
    \end{minipage}
     & 
   \begin{minipage}{.2\textwidth}
   \includegraphics[width=2.5cm,height=2.5cm]{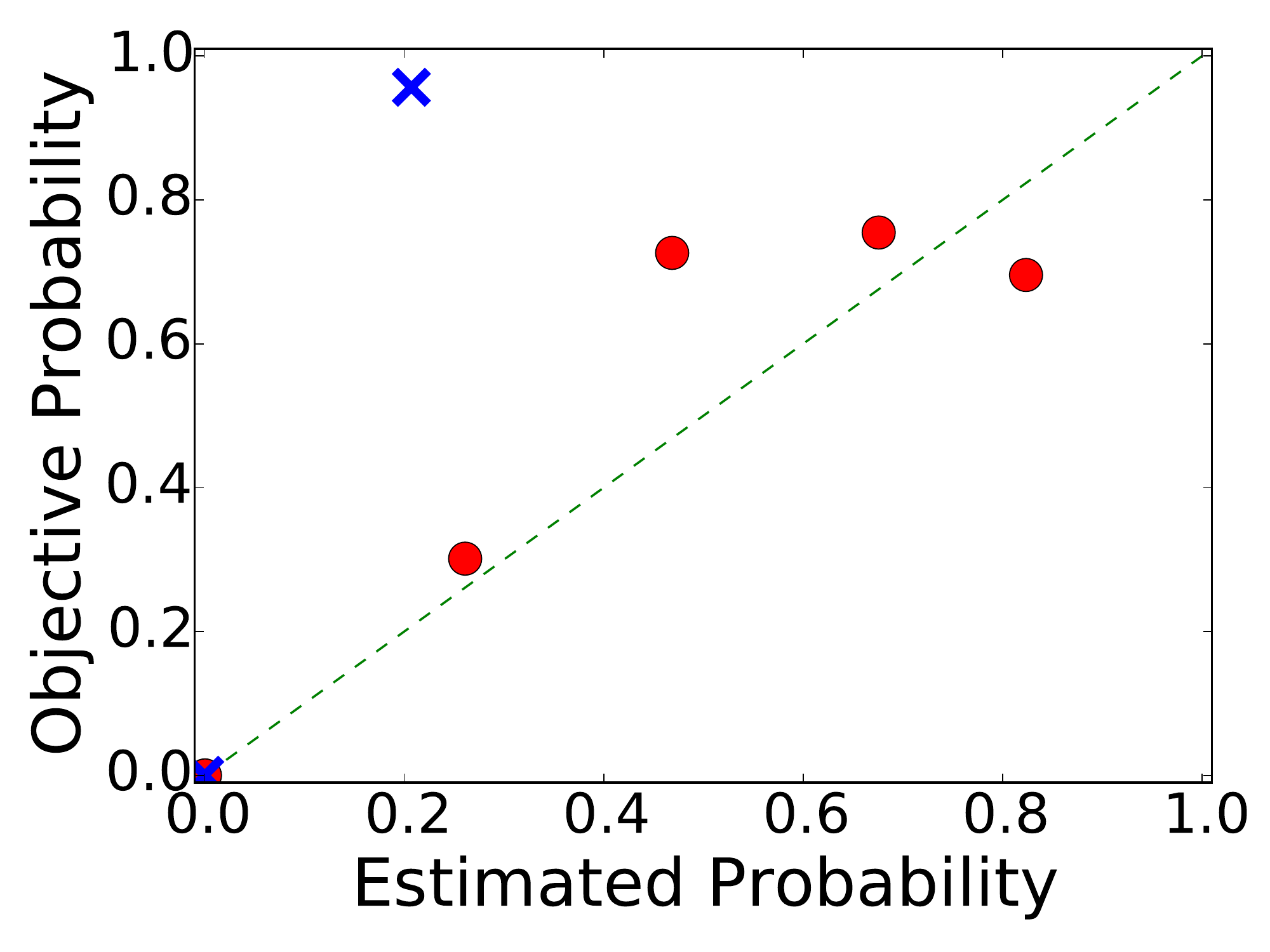}
    \end{minipage}
    &
   \begin{minipage}{.2\textwidth}
   \includegraphics[width=2.5cm,height=2.5cm]{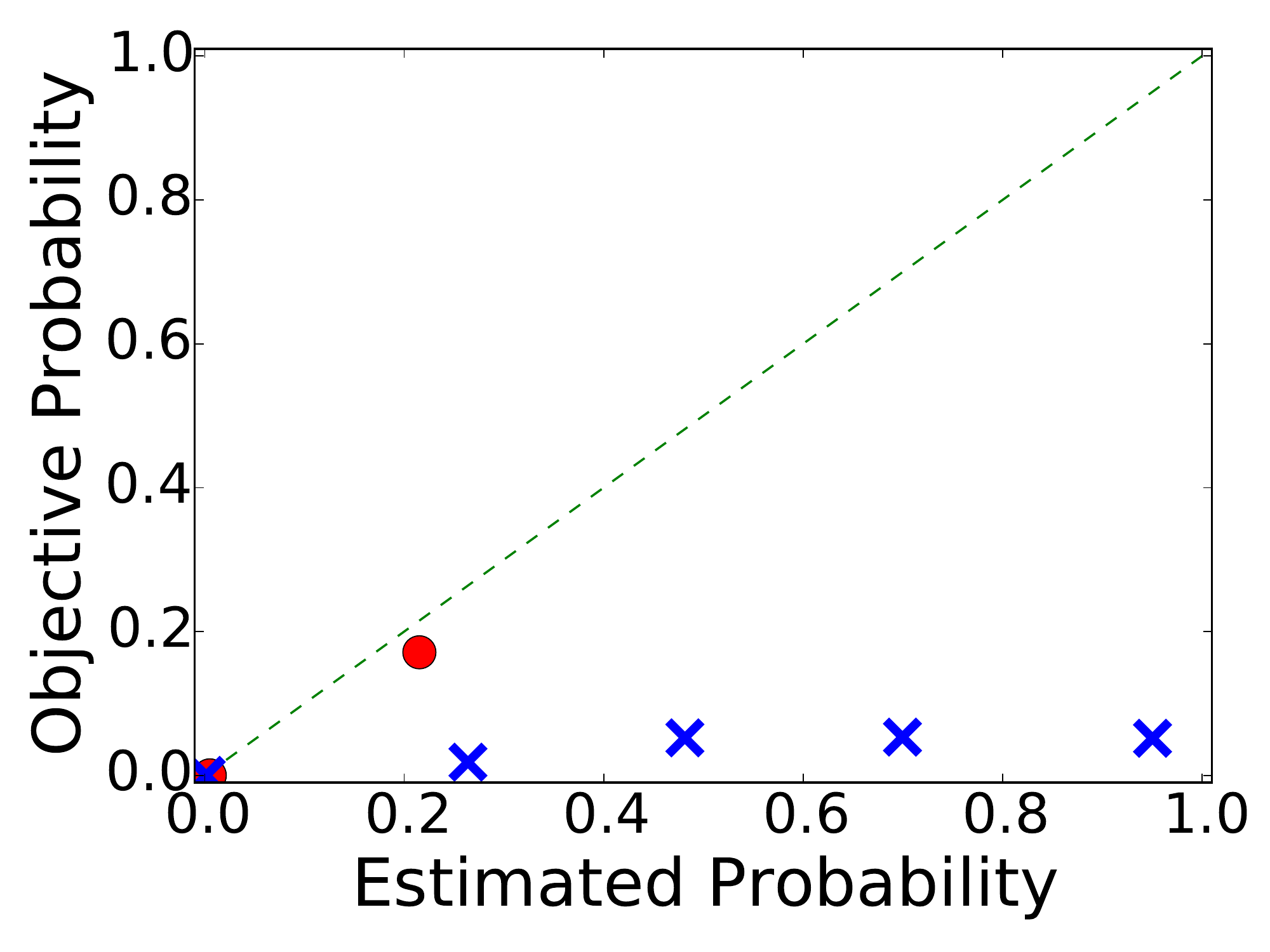}
    \end{minipage}
    &
    \begin{minipage}{.2\textwidth}
   \includegraphics[width=2.5cm,height=2.5cm]{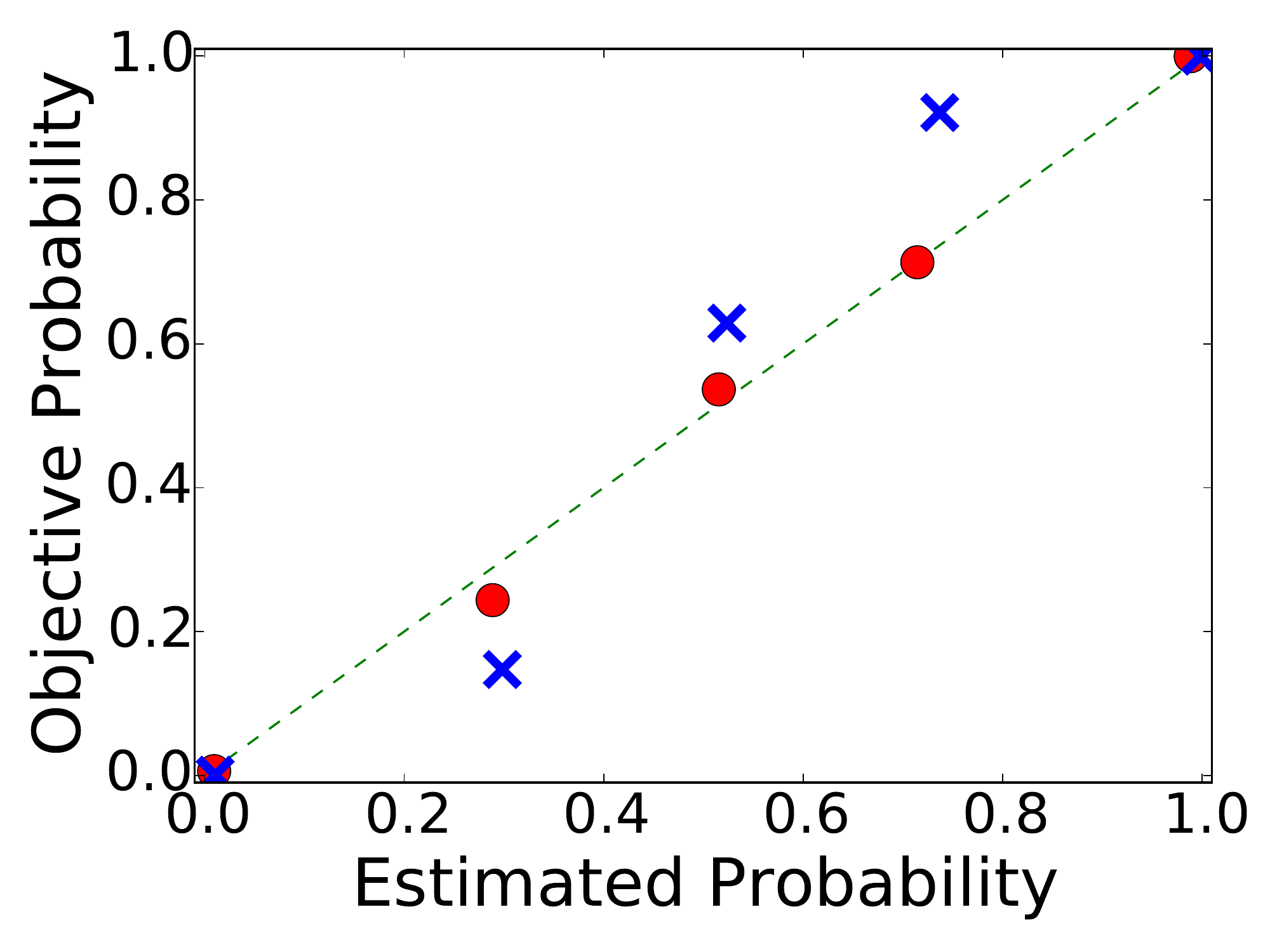}
    \end{minipage}
\vspace{-2cm}
    \\ 
    \small{V = 2k, E = 4k}
&
\begin{minipage}{.2\textwidth}
   \includegraphics[width=2.5cm,height=2.5cm]{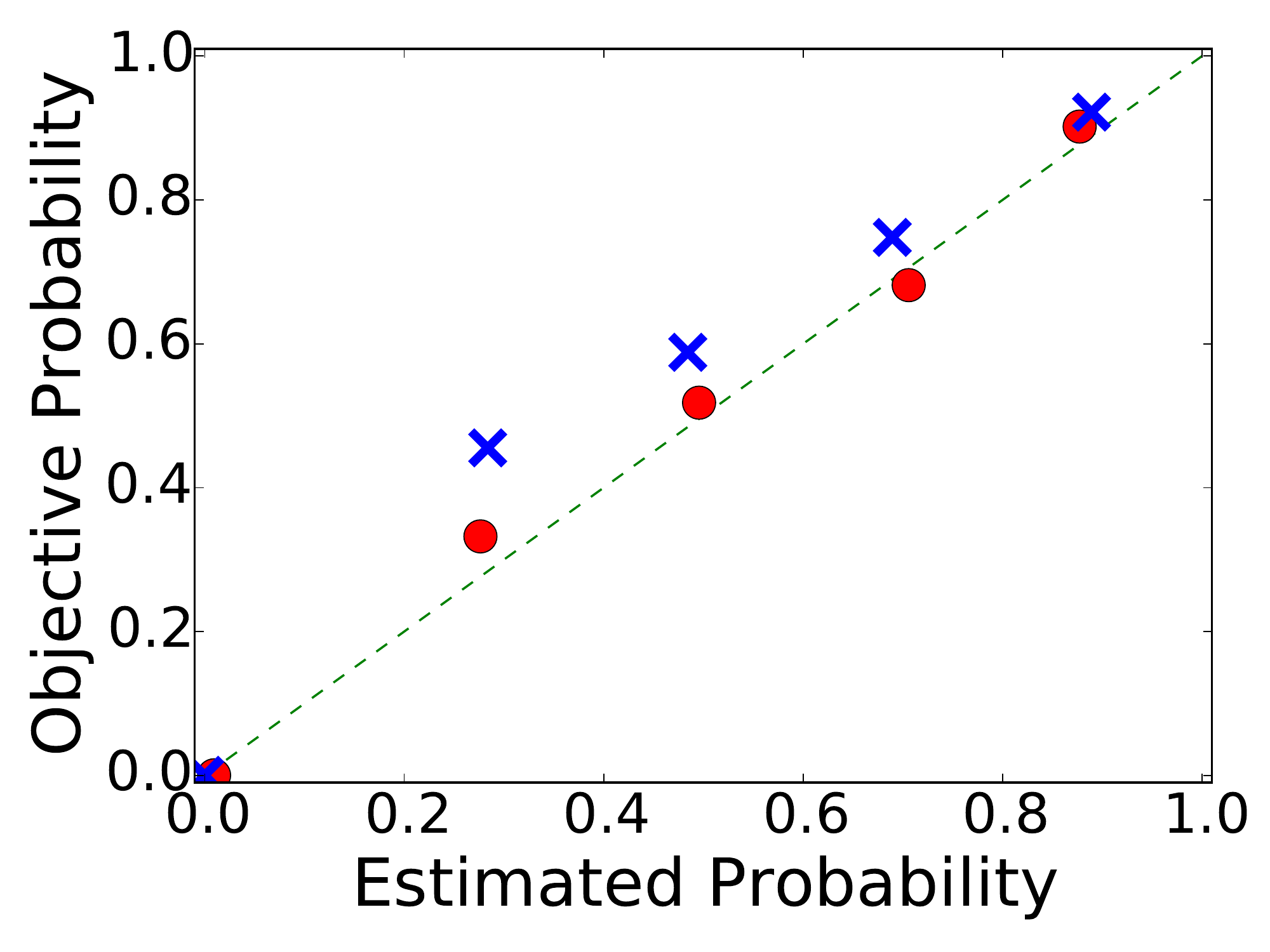}
    \end{minipage}
&
 \begin{minipage}{.2\textwidth}
\includegraphics[width=2.5cm,height=2.5cm]{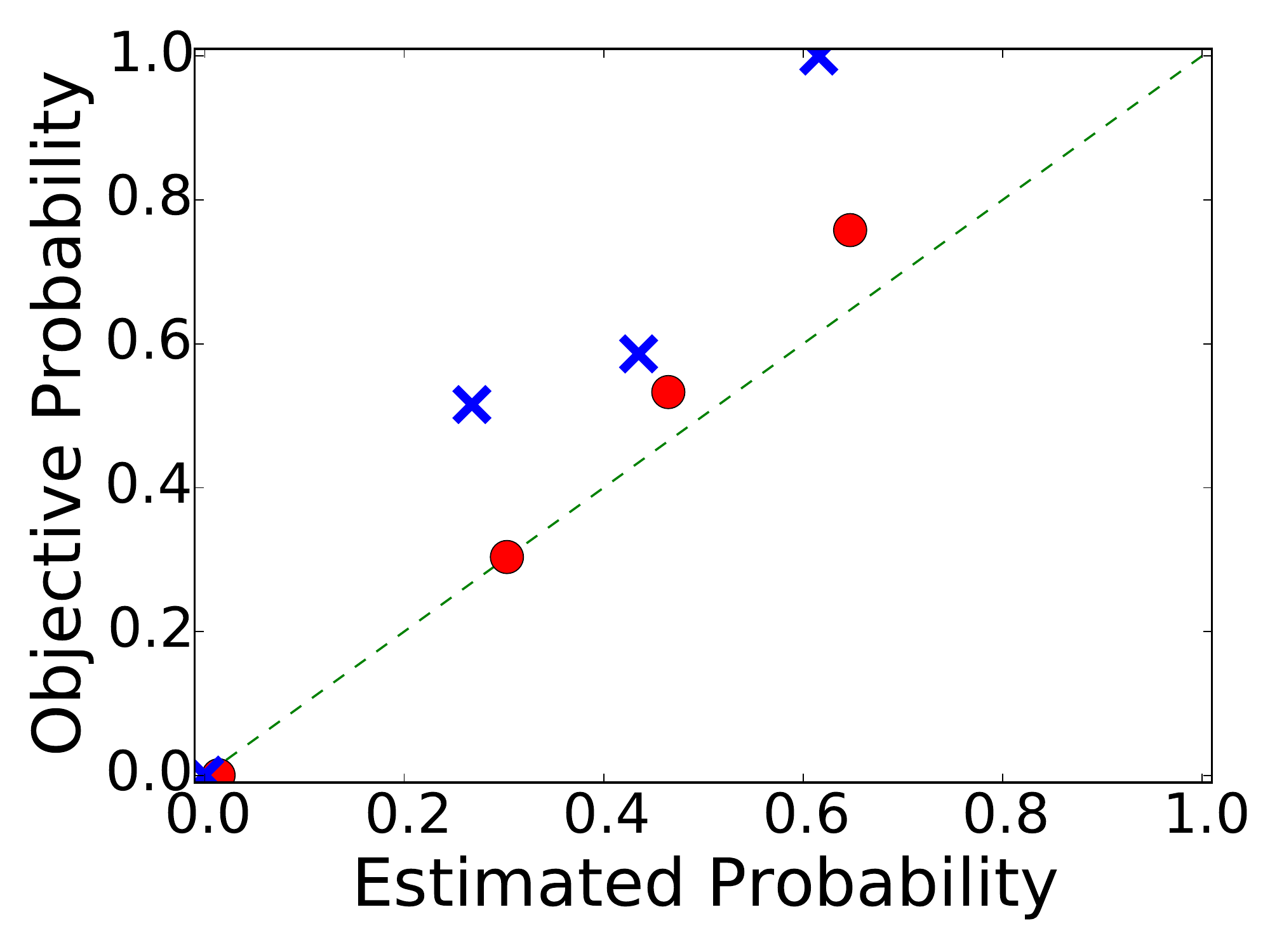}
    \end{minipage}
&
    \begin{minipage}{.2\textwidth}
\includegraphics[width=2.5cm,height=2.5cm]{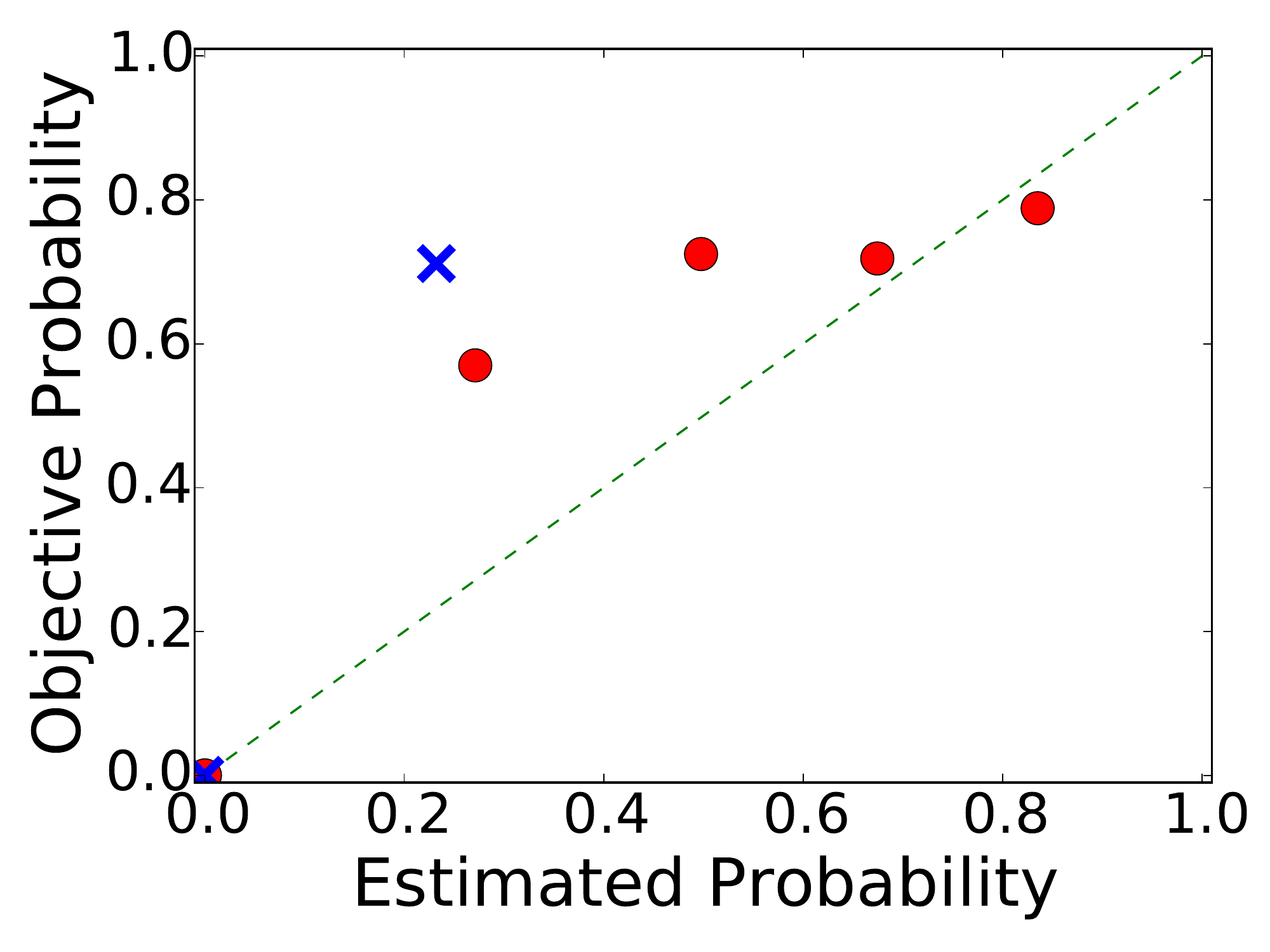}
    \end{minipage}
&
\begin{minipage}{.2\textwidth}
\includegraphics[width=2.5cm,height=2.5cm]{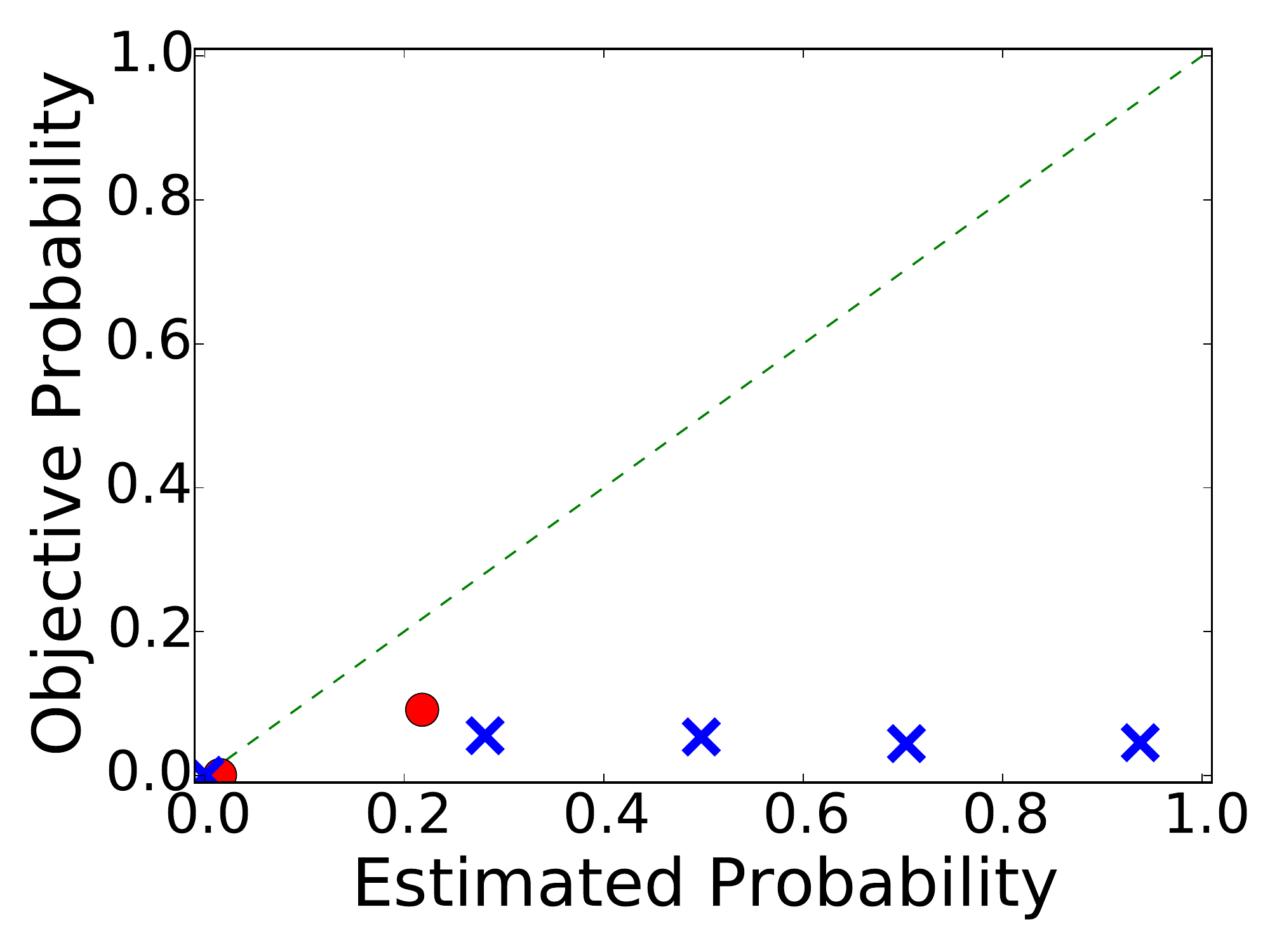}
    \end{minipage}
    &
 \begin{minipage}{.2\textwidth}
  \includegraphics[width=2.5cm,height=2.5cm]{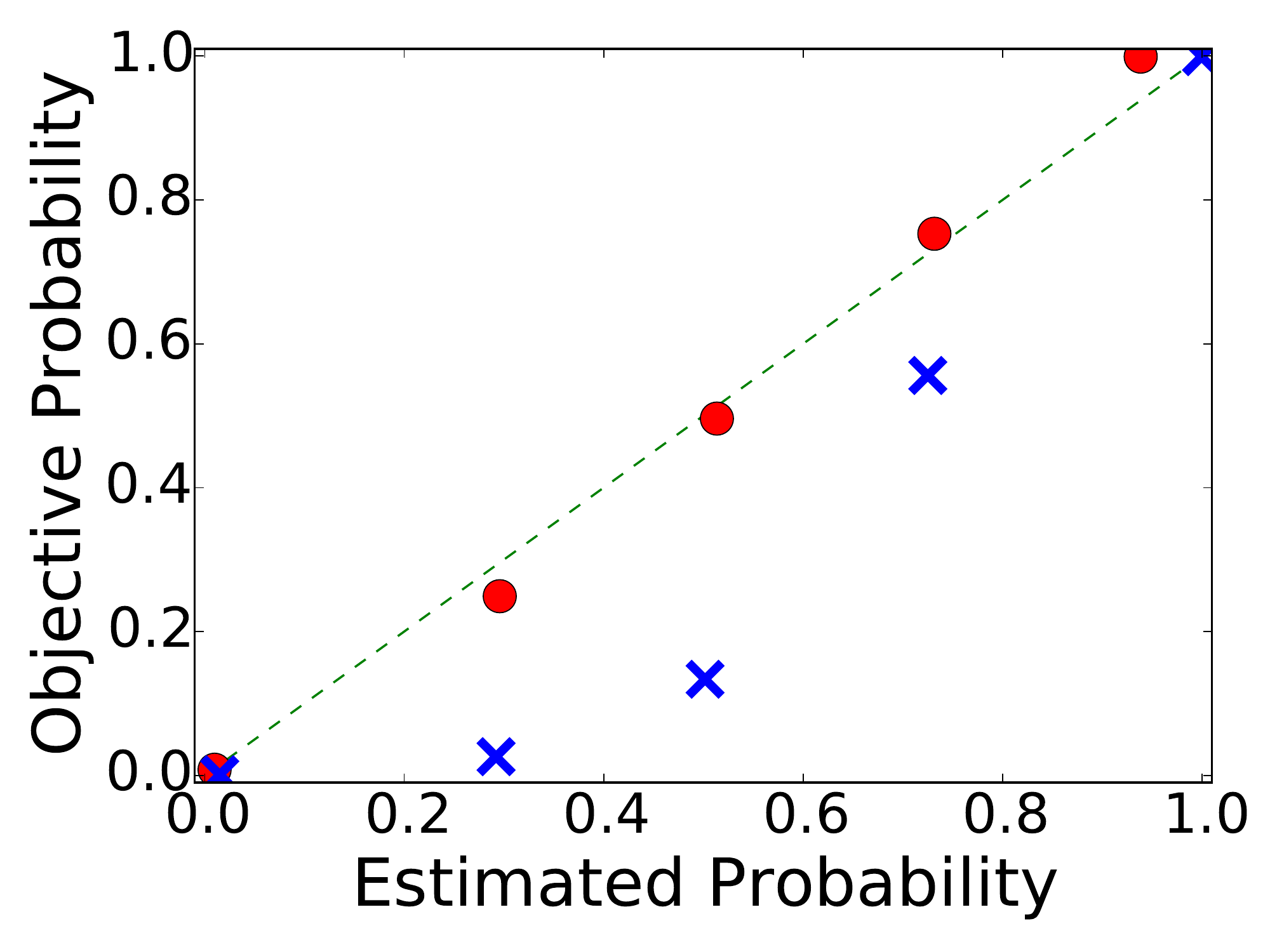}
  \end{minipage}
    \\
\end{tabular}
  }
  \captionof{figure}{\small{The calibration curves of the edge-type probabilities before (blue crosses) and after (red circles) calibration. The closer the predictions to the diagonal, the more calibrated are the probabilities. In these results, the calibration training set size is 210, the percentage of hidden variables is 0.1, and the significance level of the test of independence in RFCI is 0.005}}
  \label{fig:calibration_curve}
\vspace{-2cm}
\end{table}

Figure \ref{fig:calibration_curve} shows the proposed shallow neural network post-processing method often improves the calibration performance of the predictions for
the $A \rightarrow B$ edge-type, which is the most important edge type since it is the one that is most likely to drive experimentation. In particular, directed edges that a scientist considers to be high probability, as well as novel and important, would be prime candidates for experimental validation. Furthermore, for this edge-type, the high probability region is arguably the most critical one for making decisions about which directed arcs to investigate, such that false positive experimental investigations are minimized.

Also, the associated diagrams of the no-edge type (i.e., $A \cdots B$) in Figure \ref{fig:calibration_curve} show that the estimated probabilities are pretty well-calibrated after calibration. This is an interesting observation considering the fact that the precision and recall are also always very close to 1 for this edge type after using the post-processing calibration method. These results indicate that when the calibrated probabilities indicate with high probability that there is no edge between a pair of variables, then those nodes rarely are directly causally related. This result provides confidence in not prioritizing the experimental investigation of such node pairs for direct causal relationships.

Another interesting observation in Figure \ref{fig:calibration_curve} is that the bootstrap probabilities of the $A \leftrightarrow B$ edge-type are highly overestimated. This results in high false positive rate for that edge-type, and consequently, increases the false negative rate for other edge types. Note that post-processing the bootstrap probabilities does not generate high probabilities for the $A \leftrightarrow B$ edge-type and consequently there is no red circle in the high-probability bins but some in low probability bins, which is appropriate, because those edges, which are output by RFCI, are seldom correct.

Overall, the calibration diagrams in Figure \ref{fig:calibration_curve} show that post-processing the bootstrap probabilities using the proposed shallow-neural network model generally improves the calibration performance of the predictions. A key advantage of the shallow neural network approach for post-processing the estimated probabilities is that we can readily condition on other types of features for learning a calibration mapping (e.g., features extracted from the structure of the predicted PAGs by the RFCI method, such as the indegree of $B$ when we are generating a calibrated probability for the edge type $A \rightarrow B$). Such conditioning on local or global features of the learned graph could potentially yield improvements in the post-processed calibrated probabilities. This is an area for future research.

\section{Conclusion}
\label{sec:conclusion}
\vspace{-3mm}
In this paper, we introduced a new approach for improving the calibration of CBN structure discovery. We used a bootstrapping method to obtain estimated probabilities of the causal relationships between each pair of random variables. Although we applied the bootstrapping method to the RFCI algorithm, it can be applied with any other type of the network discovery method, as long as the method is sufficiently fast to run hundreds of times on a dataset to obtain bootstrap probability estimates. To calibrate the bootstrap probabilities, we devised a natural extension of Platt's calibration method that supports multi-class calibration using a shallow neural network. Our experiments on a wide range of large simulated datasets show that by using only a small set of instances as gold standards for training the calibration model, we can obtain substantial improvements in terms of precision, recall, and calibration, relative to the bootstrap probabilities. In future work, we plan to expand the range of simulated experiments we perform, as well as evaluate the method using real biomedical data for which the truth status is known from the literature for a relatively small subset of variables.

\textbf{Acknowledgement} We thank the members of the Center for Causal Discovery for their helpful feedback. The research reported in this publication was supported by grant U54HG008540 awarded by the National Human Genome Research Institute through funds provided by the trans-NIH Big Data to Knowledge (BD2K) initiative. The content is solely the responsibility of the authors and does not necessarily represent the official views of the National Institutes of Health. This material is also based upon work supported by the National Science Foundation under Grant No. IIS-1636786. Any opinions, findings, and conclusions or recommendations expressed in this material are those of the author(s) and do not necessarily reflect the views of the National Science Foundation.
\vspace{-3mm}

\bibliographystyle{plain}
\setlength{\bibsep}{0pt}
\scriptsize{\bibliography{nips_2017}}

\end{document}